\documentclass{article} 
\usepackage{iclr2025_conference,times}

\usepackage{amsmath,amsfonts,bm}

\def\eqref#1{equation~\ref{#1}}

\def\1{\bm{1}}

\DeclareMathAlphabet{\mathsfit}{\encodingdefault}{\sfdefault}{m}{sl}
\SetMathAlphabet{\mathsfit}{bold}{\encodingdefault}{\sfdefault}{bx}{n}

\usepackage{hyperref}
\usepackage{url}
\usepackage{microtype}
\usepackage{graphicx}
\usepackage{subcaption}
\usepackage{flafter}
\usepackage{wrapfig}
\usepackage{capt-of} 
\usepackage{booktabs} 
\usepackage{mathtools} 
\usepackage{wrapfig}   
\usepackage{placeins}  
\usepackage{dblfloatfix} %
\usepackage{adjustbox}
\usepackage{wrapfig}

\title{Hierarchy-Guided Topology Latent Flow for Molecular Graph Generation}

\author{Urvi Awasthi \& Alexander Lobo \& Leonid Zhukov\\
BCG X AI Science Institute\\
\texttt{\{awasthi.urvi, lobo.alexander, zhukov.leonid\}@bcg.com} \\
}

\iclrfinalcopy 
\begin{document}

\maketitle

\begin{abstract}

Generating chemically valid 3D molecules is hindered by \textbf{discrete bond topology}: small local bond errors can cause global failures (valence violations, disconnections, implausible rings), especially for drug-like molecules with long-range constraints. Many unconditional 3D generators emphasize coordinates and then infer bonds or rely on post-processing, leaving topology feasibility weakly controlled. We propose \textbf{Hierarchy-Guided Latent Topology Flow (HLTF)}, a planner–executor model that generates bond graphs \emph{with} 3D coordinates, using a latent multi-scale plan for global context and a constraint-aware sampler to suppress topology-driven failures. On \textbf{QM9}, HLTF achieves \textbf{98.8\% atom stability} and \textbf{92.9\% valid-and-unique}, improving \textbf{PoseBusters validity to 94.0\%} (\textbf{+0.9} over the strongest reported baseline). On \textbf{GEOM-DRUGS}, HLTF attains \textbf{85.5\%/85.0\%} validity/valid–unique–novel without post-processing and \textbf{92.2\%/91.2\%} after standardized relaxation, within \textbf{0.9} points of the best post-processed baseline. Explicit topology generation also reduces ``false-valid’’ samples that pass RDKit sanitization but fail stricter checks.\end{abstract}

\section{Introduction}
Generating chemically valid 3D molecular structures requires getting both geometry and bond topology right. A key obstacle in unconditional 3D generation is that topology is discrete and globally constrained: small local bond mistakes can cascade into valence violations, disconnected components, or implausible ring patterns, especially for drug-like molecules with long-range dependencies. Many unconditional 3D generators prioritize coordinates and then infer bonds or rely on post-processing, making topology feasibility weakly controlled and difficult to attribute.

We address this gap by \emph{explicitly} generating bond topology together with 3D coordinates. We propose \textbf{Hierarchy-Guided Latent Topology Flow (HLTF)}, a planner--executor framework that (i) evolves bond logits with feasibility-preserving continuous-time categorical dynamics, (ii) conditions topology decisions on a latent multiscale hierarchy plan that supplies long-range context, and (iii) uses an energy-regularized, annealed ODE sampler to steer sampling away from topology-driven failure modes.

We evaluate HLTF on \textsc{QM9} \cite{Ramakrishnan2014QM9} and \textsc{GEOM-DRUGS} \cite{Axelrod2022GEOM}. On \textsc{QM9}, HLTF achieves high stability and strong valid-and-unique rates, and improves plausibility under stricter validation beyond RDKit sanitization (PoseBusters \cite{Buttenschoen_2024}). On \textsc{GEOM-DRUGS}, HLTF attains strong feasibility without post-processing and remains competitive after standardized relaxation, indicating gains that are not explained solely by downstream geometry cleanup.

Our main contributions are: (1) \textbf{Planner--executor topology generation:} a coupled continuous-time formulation where a latent hierarchy plan evolves jointly with bond topology, providing global context for discrete decisions. (2) \textbf{Hierarchy-conditioned prediction:} leaf-anchored planning with sparse ancestor-masked conditioning, augmented by a lightweight hyperbolic distance signal for attention bias and pairwise features (without requiring full hyperbolic-space generation). (3) \textbf{Constraint-aware sampling:} a logit-space ODE sampler that combines endpoint prediction with modest annealed energy guidance to suppress valence/connectivity violations while preserving diverse samples.

\section{Related Work}

\textbf{\textbf{Unconditional 3D generation: validity saturation highlights topology as the bottleneck.}}
On standard benchmarks (QM9, GEOM-DRUGS), many unconditional 3D generators now report similar headline validity/stability, with gains often incremental and sometimes within run-to-run variation \cite{buttenschoen2025evaluation}. As a result, progress is increasingly driven by addressing specific global failure modes---valence violations, disconnected components, and implausible rings---that persist even when local predictions are accurate. This shifts attention from improving RDKit sanitization rates to mechanisms that provide \emph{global} structural control and to evaluation that exposes ``false-valid'' samples under stricter checks \cite{buttenschoen2025evaluation}.

\textbf{\textbf{Discrete topology modeling via flows/diffusion improves local structure but can miss long-range constraints.}}
Because atom/bond types are discrete, continuous relaxations can blur combinatorial constraints and yield globally invalid graphs. Discrete normalizing flows such as GraphDF \cite{luo2021graphdf} and categorical transport/flow-matching objectives such as CatFlow/VFM \cite{eijkelboom2024vfm} offer principled training recipes for discrete data. Recent 3D methods like SemlaFlow \cite{irwin2025semlaflow} generate coordinates jointly with discrete structure, while diffusion variants such as GruM \cite{jo2024grum} explore alternative bridge-based dynamics. However, even with strong endpoint prediction, satisfying coupled global constraints (e.g., multi-ring scaffolds and connectivity) remains challenging, motivating methods that inject explicit global context and sampling-time mechanisms to suppress topology-driven failures.

\textbf{\textbf{Hierarchy and structured inductive bias: useful global context, but often rigid or decoupled from discrete execution.}}
Hierarchical generators assemble molecules from motifs to better capture long-range dependencies \cite{jin2020motifs}, and hierarchical flows (MolGrow, MolHF) provide multi-scale invertible generation \cite{kuznetsov2021molgrow,zhu2023molhf}. Latent-space formulations and synthetic geometry similarly aim to simplify generation before decoding back to discrete topology (Pombala et al., 2025; Ketata et al., 2025), and SemlaFlow emphasizes efficient scalable 3D generation \cite{irwin2025semlaflow}. Yet these approaches can still leave feasibility to a difficult final discrete recovery step. In parallel, hyperbolic diffusion and adversarial hyperbolic autoencoding use non-Euclidean structure to represent hierarchy \cite{wen2023hgdm,fu2024hypdiff,qu2024haegan}, and tree-structured attention shows that explicitly biasing attention along hierarchical relations can improve long-range consistency \cite{nguyen2020treestructured}. Together, these lines suggest that hierarchy is valuable as a \emph{planning signal}; the remaining gap is coupling such global structure to discrete topology generation in a way that improves feasibility without requiring full generation in hyperbolic space.

\textbf{\textbf{Evaluation: stricter validity criteria are needed to expose global topology errors.}}
Prior work indicates that stronger objectives and geometric/latent representations improve sample quality, but drug-like validity remains sensitive to long-range topological constraints. This motivates reporting metrics that directly reflect dominant failure modes (validity, connectivity, ring plausibility) and complementing RDKit sanitization with stricter validators \cite{buttenschoen2025evaluation}. In 3D, SemlaFlow also argues that common evaluation can miss important physical issues and proposes energy/strain-based criteria \cite{irwin2025semlaflow}, further underscoring the need to diagnose whether improvements come from true topology feasibility versus downstream geometric cleanup.
\section{Methods}
\label{sec:methods}

\subsection{Overview}
\label{sec:methods_overview}
We propose \textbf{HLTF}, a hierarchical latent flow model for unconditional 3D molecular generation.
The method has three coupled components:
(i) a \emph{latent hierarchy plan} that encodes multi-scale structure,
(ii) a \emph{topology executor} that predicts bond types conditioned on the hierarchy,
and (iii) an \emph{E(3)-equivariant geometry predictor} for 3D coordinates.
Sampling is performed by integrating a coupled ODE in logit space (for categorical variables) and Euclidean space (for coordinates), with annealed energy guidance to encourage chemical and geometric validity.

\begin{figure*}[t]
    \centering    \includegraphics[width=0.95\textwidth]{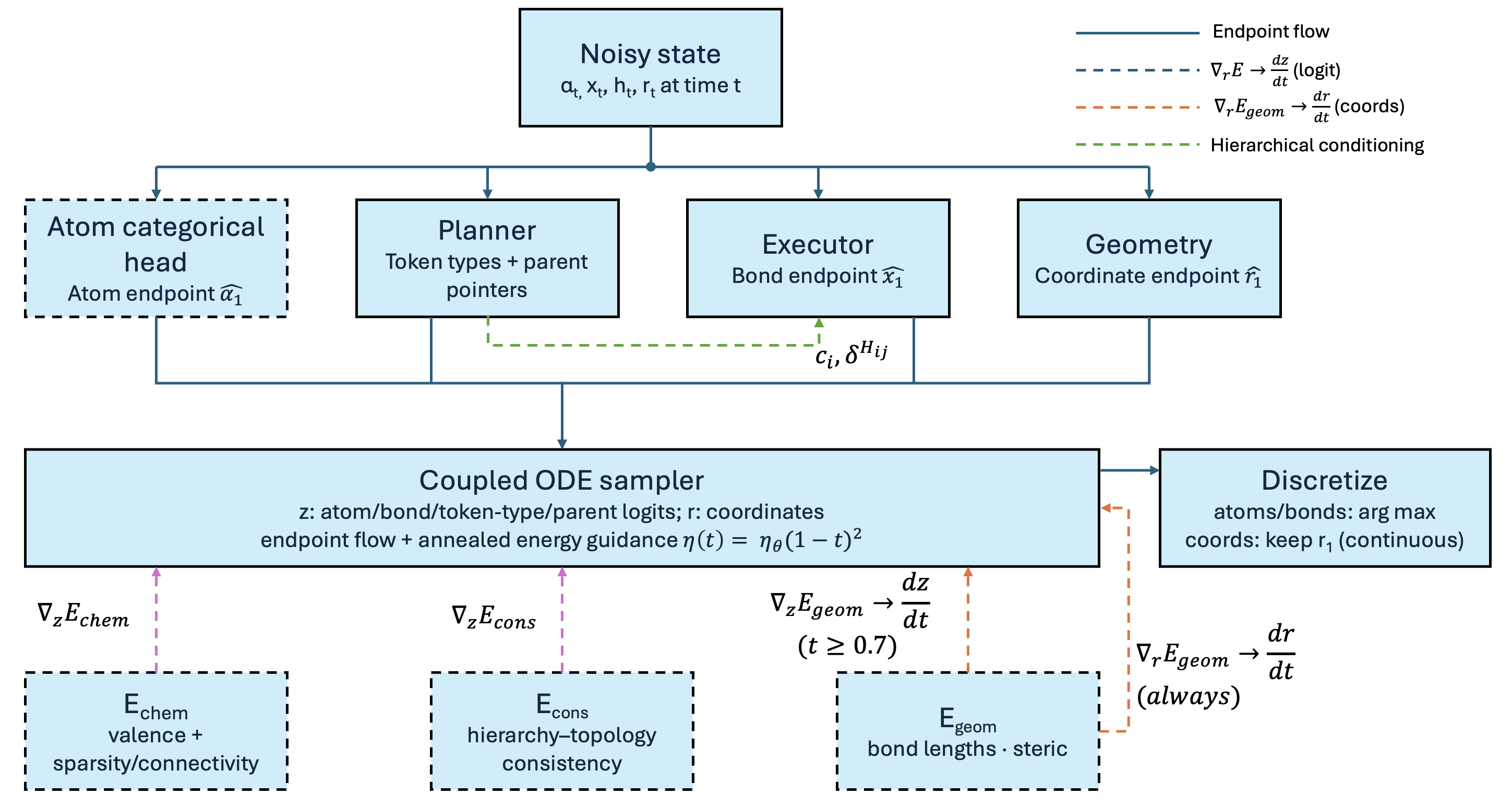}
    \caption{HLTF: Planner--Executor--Geometry Pipeline.}
    \label{fig:hltf_pipeline}
\end{figure*}

\textbf{Key contributions.}
HLTF is distinguished by:
(1) a \emph{leaf-anchored} latent hierarchy plan with probabilistic parent pointers and a soft ancestor mask enabling sparse hierarchy conditioning;
(2) a \emph{hyperbolic geometry} over hierarchy tokens used only as a lightweight distance signal (attention bias and pairwise feature), avoiding full diffusion in hyperbolic space; and
(3) a \emph{logit-space} ODE sampler that combines endpoint prediction with annealed energy guidance.

\subsection{Generative object and relaxed state}
\label{sec:relaxed_state}
Let $C_a$ be the atom-type vocabulary size and $K$ the bond-type vocabulary size (including the ``no bond'' label).
A molecule is represented as $(N,a,b,r)$ where $N$ is the number of atoms,
$a_i \in \{1,\dots,C_a\}$ is the element of atom $i$,
$b_{ij} \in \{0,\dots,K-1\}$ is the bond type for each unordered pair $(i,j)$ with $0$ indicating no bond,
and $r \in \mathbb{R}^{N\times 3}$ are atom coordinates.

HLTF maintains continuous relaxations for categorical variables during training and sampling.
Atom types use probability vectors $\alpha^i \in \Delta^{C_a-1}$ and bond types use $x^{(ij)} \in \Delta^{K-1}$,
interpreted as $\alpha_c^i \approx \mathbb{P}(a_i=c)$ and $x^{(ij)}_k \approx \mathbb{P}(b_{ij}=k)$.

\textbf{Soft bond order and soft degree.}
Associate each bond type $k$ with bond order $\omega_k$ (e.g.\ $\omega_0=0$ for no bond, $\omega_1=1$ single, $\omega_2=2$ double, etc.).
The expected bond order for pair $(i,j)$ is $\langle \omega, x^{(ij)}\rangle = \sum_{k=0}^{K-1}\omega_k x^{(ij)}_k$,
and the soft degree (soft valence) of atom $i$ is
\begin{equation}
\deg_i(x) \;=\; \sum_{j\neq i}\langle \omega, x^{(ij)}\rangle.
\end{equation}

\textbf{Coordinate gauge.}
We enforce translation invariance by recentering coordinates after each ODE step so that $\sum_{i=1}^N r_i = \mathbf{0}$.
Rotational symmetry is handled by an E(3)-equivariant geometry predictor and random global rotations during training.

\subsection{Latent hierarchy plan with leaf anchoring}

\label{sec:plan_state_main}
HLTF introduces a latent \emph{hierarchy plan} $h_t$ represented as a rooted token tree.
Each molecule has:
(i) a ROOT token,
(ii) $M$ \emph{motif tokens} representing multi-atom substructures, and
(iii) $N$ \emph{atom-leaf tokens} $\{\ell(1),\dots,\ell(N)\}$ with a fixed one-to-one correspondence between atoms and leaves.

\textbf{Leaf anchoring.}
Atom $i$ always connects to the hierarchy through its dedicated leaf token $\ell(i)$.
This removes any test-time atom$\rightarrow$token alignment problem and enables sparse conditioning: atom $i$ only needs to attend to tokens on its ancestor chain from $\ell(i)$ to ROOT.

\textbf{Deterministic hierarchy builder.}
At training time, we extract $h_1$ using a deterministic builder. For QM9, we use BRICS-based fragmentation \cite{Degen2008DrugLikeFragmentsBRICS}. For GEOM-DRUGS, we use a \emph{ring-first hybrid strategy}: extract fused ring systems (merging rings sharing $\geq 1$ atom), apply RECAP fragmentation~\citep{lewell1998retrosynthetic} to acyclic regions (more conservative than BRICS, preserving drug scaffolds), and merge smallest motifs if count exceeds $M_{\max}$. See Appendix~A for details.

\textbf{Plan variables.}
Each token $\alpha$ has a categorical type distribution $y_t^\alpha \in \Delta^{C_h-1}$ over a token-type vocabulary of size $C_h$.
For each non-root token $\alpha > 1$, we maintain a distribution over parents
$\boldsymbol{\rho}_t^\alpha \in \Delta^{A_{\max}-1}$, where $A_{\max}=1+M_{\max}+N_{\max}$ is the token budget used for padding.
We enforce a single root (token $\alpha=1$) and prevent cycles by a causal ordering constraint:
\begin{equation}
\rho_t^\alpha[\beta] = 0 \quad \text{for } \beta \ge \alpha \quad (\alpha>1),
\end{equation}
followed by renormalization over valid parents $\beta < \alpha$.

\textbf{Soft ancestor mask.}
Because parent pointers are probabilistic during generation, we compute a \emph{soft} ancestor probability
$\pi_{i\alpha}(h_t)\in[0,1]$, the probability that token $\alpha$ lies on the ancestor chain of leaf $\ell(i)$.
Using the causal ordering, these probabilities are computed efficiently via dynamic programming:
\begin{align}
\pi_{i,\ell(i)} &= 1, \label{eq:ancestor_base_main}\\
\pi_{i,\beta} &= \sum_{\alpha > \beta} \pi_{i,\alpha}\,\rho_t^\alpha[\beta], \qquad \beta < \ell(i), \label{eq:ancestor_recursive_main}\\
\pi_{i,\beta} &= 0, \qquad \beta > \ell(i). \label{eq:ancestor_future_main}
\end{align}
We use $\pi_{i\alpha}$ as a soft attention mask for hierarchy conditioning (Section~\ref{sec:hier_conditioning}).

\begin{wrapfigure}{r}{0.42\textwidth}
    \centering
    \vspace{-0.8em}
    \includegraphics[width=\linewidth]{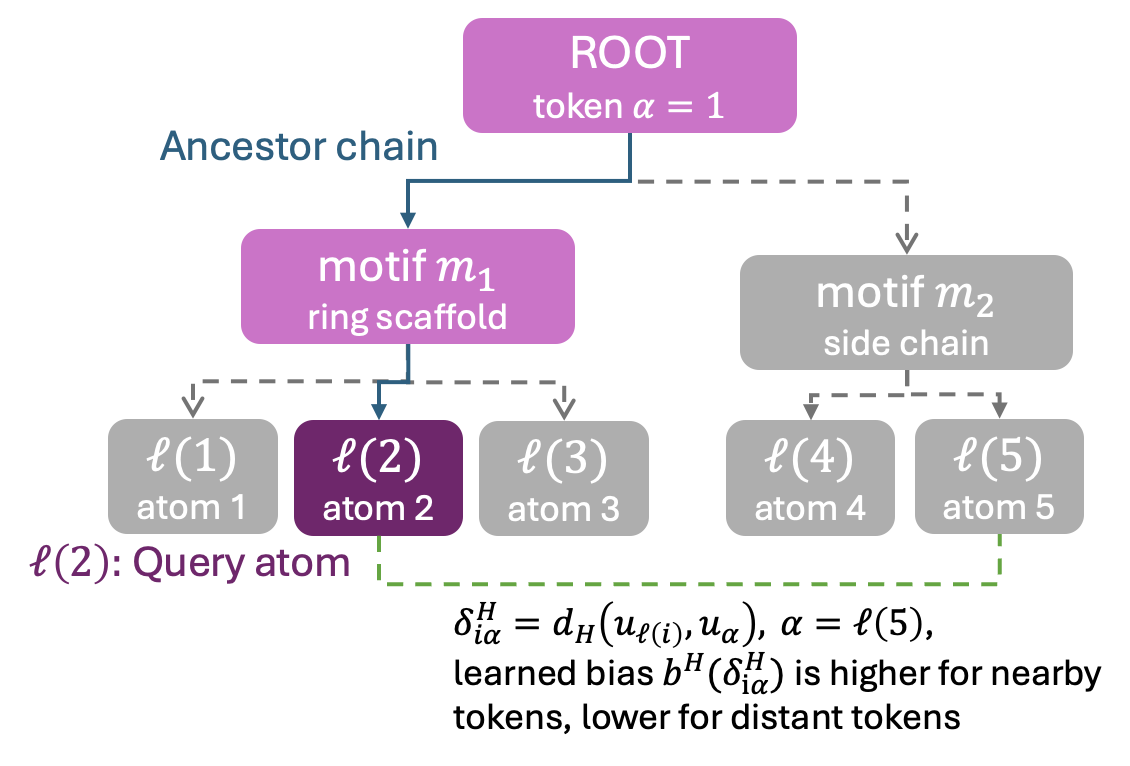}
    \caption{Leaf-anchored hierarchy conditioning.}
    \label{fig:leaf_anchor_mask}
    \vspace{-1.0em}
\end{wrapfigure}

\subsection{Hierarchy conditioning with hyperbolic token geometry}
\label{sec:hier_conditioning}
We encode hierarchy tokens into Euclidean states and additionally assign each token a hyperbolic coordinate used only as a distance signal.

\textbf{Euclidean hierarchy encoder.}
Let $E_y$ be a learned token-type embedding matrix.
For token $\alpha$, we embed its uncertain type distribution by the expected embedding
\begin{equation}
e_\alpha \;=\; E_y^\top y_t^\alpha.
\end{equation}
We refine $\{e_\alpha\}$ into Euclidean token states $\{h_\alpha\in\mathbb{R}^H\}$ via $L_h$ layers of message passing on the (probabilistic) tree,
weighting messages from child $\alpha$ to candidate parent $\beta$ by $\rho_t^\alpha[\beta]$.
We then produce attention keys/values $k_\alpha = W_k h_\alpha$ and $v_\alpha=W_v h_\alpha$.

\textbf{Hyperbolic token geometry (distance signal).}
We map each Euclidean token state to a hyperbolic coordinate $u_\alpha \in \mathcal{B}_c^{d_H}$ (Poincar\'e ball with curvature $c>0$)\footnote{See Appendix \ref{app_hyp_hier_geometry}}
via an exponential map from the origin:
\begin{align}
\tilde{u}_\alpha &= W_H h_\alpha, \nonumber\\
u_\alpha &= \exp_0^c(\tilde{u}_\alpha)
= \frac{\tanh(\sqrt{c}\|\tilde{u}_\alpha\|)}{\sqrt{c}\|\tilde{u}_\alpha\|}\tilde{u}_\alpha.
\end{align}
We compute hyperbolic distances $d_H^c(u,v)$ using the standard Poincar\'e metric \cite{nickel2017poincareembeddingslearninghierarchical}.
For atom $i$ and token $\alpha$, define the hierarchy distance
\begin{equation}
\delta_{i\alpha}^H \;=\; d_H^c(u_{\ell(i)}, u_\alpha),
\label{eq:hyperbolic_signal}
\end{equation}
and for atom pairs $(i,j)$ define $\delta_{ij}^H = d_H^c(u_{\ell(i)}, u_{\ell(j)})$.

\textbf{Sparse masked atom$\rightarrow$hierarchy attention.}
Each atom $i$ attends to hierarchy tokens with an additive bias from hyperbolic distance and a soft ancestor mask:
\begin{align}
\ell_{i\alpha} &= \frac{q_i^\top k_\alpha}{\sqrt{d}} + b_H(\delta_{i\alpha}^H)
+ \log(\pi_{i\alpha}(h_t)+\epsilon_m), \label{eq:attn_scores_main}\\
w_{i\alpha} &= \mathrm{softmax}_\alpha(\ell_{i\alpha}), \quad
c_i = \sum_{\alpha} w_{i\alpha} v_\alpha, \label{eq:attn_context_main}
\end{align}
where $b_H(\cdot)$ is a small learned MLP, $q_i$ is an atom query vector derived from the atom backbone state,
and $\epsilon_m$ is a small constant for numerical stability.
The resulting hierarchy context $c_i$ is used for bond prediction.

\subsection{Planner--Executor--Geometry predictors}
\label{sec:predictors}
HLTF uses three endpoint predictors that take the current relaxed state $(\alpha_t,x_t,h_t,r_t,t)$ and predict the $t{=}1$ endpoints.

\textbf{Planner (hierarchy endpoint predictor).}
The planner predicts token types and parent pointers:
\begin{equation}
q_\phi(h_1 \mid \alpha_t, x_t, h_t, r_t, t) =  \left(\prod_{\alpha=1}^{A_{\max}} \mathrm{Cat}(y_1^\alpha \mid \nu_\phi^\alpha)\right) \notag \times \left(\prod_{\alpha=2}^{A_{\max}} \mathrm{Cat}(\mathrm{par}(\alpha) \mid \rho_\phi^\alpha)\right),
\end{equation}
 where $\nu_\phi^\alpha = \nu_\phi^\alpha(\alpha_t, x_t, h_t, r_t, t)$ are the predicted parameters for token $\alpha$'s type, and $\rho_\phi^\alpha = \rho_\phi^\alpha(\alpha_t, x_t, h_t, r_t, t)$  are the predicted parameters for token $\alpha$'s parent pointer (with the causal parent masking applied to $\rho_\phi^\alpha$). Here $\mathrm{Cat}(\cdot \mid \nu)$ denotes a categorical distribution with parameters $\nu$.

\textbf{Executor (bond endpoint predictor).}
The executor predicts the final bond topology $x_1$ by independently predicting each edge:
\begin{equation}
q_\theta(x_1 \mid \alpha_t, x_t, h_t, r_t, t)
= \prod_{i<j} \mathrm{Cat}\!\Bigl(
b_{ij} \,\Bigm|\,
\mu_\theta^{(ij)}(\alpha_t, x_t, h_t, r_t, t)
\Bigr),
\label{eq:executor}
\end{equation}
where $\mu_\theta^{(ij)}$ are the predicted categorical parameters for the bond type between atoms $i$ and $j$, and $b_{ij} \in \{0, 1, \ldots, K-1\}$ denotes the bond type.

\textbf{Geometry endpoint predictor.}
We predict final coordinates using an E(3)-equivariant network:
\begin{equation}
m_\psi(r_t,\alpha_t,x_t,h_t,t)\in\mathbb{R}^{N\times 3}.
\label{eq:geom_main}
\end{equation}

\textbf{Atom backbone and edge head (summary).}
Atoms obtain hidden states $s_i\in\mathbb{R}^H$ from an EGNN-style message-passing backbone operating on $(\alpha_t,x_t,r_t,t)$ \cite{satorras2022enequivariantgraphneural}.
For each pair $(i,j)$, we predict bond logits using an edge MLP whose activations are modulated by a small hypernetwork (FiLM),
conditioned on atom features, hierarchy contexts $(c_i,c_j)$, time $t$, geometry, and the pairwise hierarchy distance $\delta_{ij}^H$ \cite{perez2017filmvisualreasoninggeneral}.
All architecture and feature details are deferred to Appendix~\ref{app:impl_details}.

\subsection{Training via endpoint prediction}
\label{sec:training_main}
We train by sampling $t\sim\mathcal{U}(0,1)$ and constructing noisy states by linear interpolation between data endpoints
$(\alpha_1,x_1,h_1,r_1)$ and priors $(\alpha_0,x_0,h_0,r_0)$:
\begin{equation}
(\alpha_t,x_t,h_t,r_t) \;=\; t(\alpha_1,x_1,h_1,r_1) + (1-t)(\alpha_0,x_0,h_0,r_0).
\label{eq:interpolants_main}
\end{equation}
The model predicts endpoints $(\hat{\alpha}_1,\hat{x}_1,\hat{h}_1,\hat{r}_1)$ and we minimize a weighted sum of cross-entropy losses
for categorical endpoints and MSE for coordinates:
\begin{equation}
\mathcal{L}
= \mathcal{L}_{\mathrm{atom}} + \lambda_b \mathcal{L}_{\mathrm{bond}} + \lambda_h \mathcal{L}_{\mathrm{plan}} + \lambda_r \mathcal{L}_{\mathrm{coord}}.
\label{eq:loss_main}
\end{equation}
Exact priors, loss weighting choices, and all optimization hyperparameters are provided in Appendix~\ref{app:impl_details}.

\subsection{Sampling: coupled ODE in logit space and coordinates}
\label{sec:sampling_main}
Direct ODE integration in probability space can violate simplex constraints.
HLTF instead integrates categorical variables in \emph{logit space}.
Let $z$ collect all logits for atom types, bond types, token types, and parent pointers, and let $(\alpha,x,h)=\mathrm{softmax}(z)$.

\textbf{Simplex feasibility and logit-space dynamics.}
Integrating ODEs directly in probability space can violate simplex constraints: after an ODE step, probabilities may
become negative or fail to sum to $1$. To maintain feasibility throughout sampling, HLTF integrates all categorical
dynamics in \emph{logit space}.

For each categorical variable (atom types, bond types, hierarchy token types, parent pointers), we maintain unconstrained
logits $z \in \mathbb{R}^d$ and map to probability vectors via softmax: $p=\operatorname{softmax}(z)$.
Since softmax always produces valid probability distributions (non-negative, summing to $1$), the constraints are
automatically satisfied at every step regardless of the logit values.

Energy terms are defined as functions of probabilities $(\alpha,x,h)$ and coordinates $r$. Gradients of these energies
with respect to logits are computed via automatic differentiation (chain rule through the softmax), ensuring that all
updates respect the simplex constraints.

\textbf{Endpoint-logit target.}
We denote by $\mathrm{Logits}_\Theta(\alpha,x,h,r,t)$ the concatenation of all endpoint logits predicted by the model's
categorical heads. This includes pre-softmax predictions for: atom types (for each atom $i$), bond types (for each pair
$(i,j)$), hierarchy token types (for each token $\alpha$), and parent pointers (for each token $\alpha > 1$).
These predicted logits serve as the target toward which the current logit state moves during ODE integration.

Sampling integrates from $t=0$ to $t=1$ using the coupled dynamics
\begin{align}
\frac{dz}{dt}
&= \frac{\mathrm{Logits}_\Theta-z}{1-t+\varepsilon} \nonumber\\
&\quad{-}\eta_1\nabla_z E_1
{-}\eta_2\nabla_z E_2
{-}\eta_3\mathbf{1}[t{\ge}0.7]\nabla_z E_3,
\label{eq:ode_z}
\\
\frac{dr}{dt}
&= \frac{m_\psi-r}{1-t+\varepsilon}
{-}\eta_4\nabla_r E_3,
\label{eq:ode_r}
\end{align}
with $E_1{=}E_{\text{chem}}(\alpha,x)$, $E_2{=}E_{\text{cons}}(x,h)$, $E_3{=}E_{\text{geom}}(r,\alpha,x,t)$, and all $\eta$ time-dependent.

The first term in each equation implements \emph{endpoint flow}, driving the state toward the network-predicted endpoint.
The remaining terms implement \emph{annealed energy guidance} to encourage validity early in the trajectory while allowing endpoint prediction to dominate near $t=1$.
We apply geometry-to-topology guidance only after a late-time threshold $t_{\mathrm{geom}}$ to reduce stiffness when coordinates are highly noisy.
After each solver step, we re-center $r$, re-apply padding masks, and enforce the causal parent mask.

\textbf{Energy terms (summary).}
$E_{\mathrm{chem}}$ encodes soft chemical constraints (e.g.\ valence, sparsity/connectivity),
$E_{\mathrm{cons}}$ encourages hierarchy--topology consistency by aligning edge probabilities with hierarchy proximity (via hyperbolic distances),
and $E_{\mathrm{geom}}$ encourages realistic 3D structure (bond lengths and steric repulsion).
Full formulations and constants are provided in Appendix~\ref{app:energies}.

\subsection{Discretization}
\label{sec:discretization_main}
At $t\approx 1$, we discretize by $\arg\max$:
$a_i=\arg\max_c \alpha_1^i[c]$ and $b_{ij}=\arg\max_k x_1^{(ij)}[k]$.
Optionally, we apply a post-processing valence repair that deletes low-confidence incident bonds until all atoms satisfy their maximum valence;
details are in Appendix~\ref{app:impl_details}.

\section{Experiments}

\begin{table}[b]
\centering

\caption{\textbf{QM9 results.}
(a) Unconditional generation; (b) novelty and PoseBusters validity.
Baselines reproduced from GCDM Table~1 \cite{morehead2023gcdm}.
Arrows $\uparrow/\downarrow$ indicate whether higher/lower is better.
Asterisk * denotes best; underline denotes second best.}
\label{tab:qm9}

\begin{subtable}[t]{0.61\linewidth}
\centering
\caption{Unconditional generation.}
\label{tab:qm9_main}
\scriptsize
\setlength{\tabcolsep}{2.5pt}
\renewcommand{\arraystretch}{1.02}
\begin{tabular}{lcccc}
\toprule
 & \shortstack{AS\\(\%)$\uparrow$} &
\shortstack{MS\\(\%)$\uparrow$} &
\shortstack{Val\\(\%)$\uparrow$} &
\shortstack{Val\&Uniq\\(\%)$\uparrow$} \\
\midrule
E-NF & 85.0 & 4.9 & 40.2 & 39.4 \\
G-Schnet & 95.7 & 68.1 & 85.5 & 80.3 \\
GDM & 97.0 & 63.2 & -- & -- \\
GDM-aug & 97.6 & 71.6 & 90.4 & 89.5 \\
EDM & $98.7\!\pm\!0.1$ & $82.0\!\pm\!0.4$ & $91.9\!\pm\!0.5$ & $90.7\!\pm\!0.6$ \\
Bridge & $98.7\!\pm\!0.1$ & $81.8\!\pm\!0.2$ & -- & 90.2 \\
Bridge+Force & $98.8\!\pm\!0.1$ & $84.6\!\pm\!0.3$ & 92.0 & 90.7 \\
GraphLDM & 97.2 & 70.5 & 83.6 & 82.7 \\
GraphLDM-aug & 97.9 & 78.7 & 90.5 & 89.5 \\
GeoLDM & $98.9\!\pm\!0.1$* & $89.4\!\pm\!0.5$* & $93.6\!\pm\!0.2$ & $92.7\!\pm\!0.5$ \\
GCDM & $98.7\!\pm\!0.1$ & $85.7\!\pm\!0.4$ & $94.8\!\pm\!0.2$* & $93.3\!\pm\!0.0$* \\
\midrule
\textbf{HLTF (ours)} & \underline{$98.8\!\pm\!0.1$} & \underline{$86.8\!\pm\!0.3$} & \underline{$93.8\!\pm\!0.6$} & \underline{$92.9\!\pm\!0.2$}\\
\bottomrule
\end{tabular}
\end{subtable}
\hfill
\begin{subtable}[t]{0.37\linewidth}
\centering
\caption{Novelty and PoseBusters validity.}
\label{tab:qm9_pb}
\footnotesize
\setlength{\tabcolsep}{3pt}
\renewcommand{\arraystretch}{1.05}
\resizebox{\linewidth}{!}{%
\begin{tabular}{lccc}
\toprule
Metric &
\shortstack{GeoLDM} &
\shortstack{GCDM} &
\shortstack{\textbf{HLTF}\\\textbf{(ours)}} \\
\midrule
AS (\%) $\uparrow$            & $98.9\pm0.1$*      & $98.7\pm0.1$    & \underline{$98.8\pm0.1$} \\
MS (\%) $\uparrow$            & $89.4\pm0.5$*      & $85.7\pm0.4$    & \underline{$86.8\pm0.3$} \\
Val (\%) $\uparrow$           & $93.6\pm0.2$       & $94.8\pm0.2$*   & \underline{$93.8\pm0.6$} \\
Val\&Uniq (\%) $\uparrow$     & $92.7\pm0.5$       & $93.3\pm0.0$*   & \underline{$92.9\pm0.2$} \\
Novel (\%) $\uparrow$         & $53.5\pm0.6$       & $58.7\pm0.5$*   & \underline{$56\pm0.7$} \\
PB-Valid (\%) $\uparrow$      & $93.1\pm0.4$       & $91.9\pm0.5$    & \textbf{$94\pm0.5$*} \\
\bottomrule
\end{tabular}}
\end{subtable}

\end{table}

\textbf{Motivation.}
We use two benchmarks to test complementary claims: on \textbf{QM9}, whether an explicit bond-topology generator can remain competitive under standard protocols even though many protocols implicitly favor inferred-bond coordinate models; on \textbf{GEOM-DRUGS}, whether improving \emph{topology feasibility} (valence/connectivity/ring consistency) is the main driver of end-to-end success on drug-like molecules, beyond post-hoc geometry relaxation.

\subsection{Benchmarks, protocols, and metrics}
We follow the published 3D unconditional-generation protocols adopted by prior work. On QM9 we report atom stability (AS), molecule stability (MS), RDKit validity (Val), and Val\&Uniq on 10{,}000 unconditional samples. On GEOM-DRUGS we report Valid, Valid\&Unique (V\&U), and Valid\&Unique\&Novel (V\&U\&N) on 100{,}000 samples, both raw and after standardized post-processing (largest fragment, explicit hydrogens, UFF relaxation\cite{doi:10.1021/ja00051a040}).  Since our central hypothesis concerns topology feasibility (valence/connectivity/ring consistency) as the dominant bottleneck, these success-rates directly reflect the failure modes we target. While these protocols directly measure end-to-end feasibility, they do not fully characterize distribution matching (e.g., scaffold/property coverage) or conformer realism beyond pass/fail validity and optional relaxation. We therefore interpret success-rate gains primarily as reductions in topology-driven failure modes, and leave broader distributional and geometry-quality analyses to future work (see Appendix \ref{app:limitations} for related limitations).

\subsection{Evaluation modes}
We omit NLL because sampling-time energy guidance breaks exact likelihood computation.

\textbf{Explicit-topology strictness.} Many pipelines infer bonds from coordinates, which can partially absorb topology errors. HLTF generates bonds and is evaluated using its \emph{generated} topology, which is a stricter requirement because topology mistakes directly reduce Valid/V\&U/V\&U\&N; we therefore interpret comparisons to inferred-bond pipelines as conservative for HLTF. For completeness, reporting HLTF results under the same inferred-bond evaluation used by coordinate-only pipelines (and/or evaluate baselines under explicit-topology constraints) to isolate protocol effects would be informative; we leave this matched-protocol study to future work.

\textbf{Dataset-specific builders.}
We use BRICS for QM9 and a ring-first hybrid RECAP builder for GEOM-DRUGS (Sec.~3.3), which better preserves multi-ring scaffolds.

\subsection{Main training results: QM9}
\textbf{Hypothesis.}
If topology is the main bottleneck, an explicit topology+geometry model should remain competitive on stability/validity under protocol-matched evaluation, and show clearer gains under stricter plausibility checks. In Table~\ref{tab:qm9_main}, HLTF is within $0.4$ points of the best Val\&Uniq (92.9 vs.\ 93.3) and within $0.1$ of the best AS (98.8 vs.\ 98.9), despite being evaluated on generated (not inferred) bonds.

\textbf{Result} Under this protocol, coordinate-only baselines can benefit from bond perception, whereas HLTF is evaluated model-faithfully on the bonds it generates. Despite this stricter requirement, HLTF remains competitive on stability and RDKit validity, suggesting that explicit bond generation does not inherently trade off protocol metrics. This motivates evaluating plausibility with stricter validators beyond RDKit sanitization. 

\textbf{Beyond RDKit sanitization.}
Table~\ref{tab:qm9_main} reports novelty and PoseBusters validity\cite{Buttenschoen_2024}. HLTF  improves PB-Valid by $+0.9$ over the strongest baseline reported while staying within $0.5$ of the best Val\&Uniq. Higher PB-Valid at comparable RDKit validity indicates fewer ``false-valid'' samples that sanitize in RDKit but violate stricter structural constraints at comparable headline validity.

\textbf{Why PoseBusters matters here.} RDKit validity can miss chemically implausible structures that nevertheless sanitize. PoseBusters applies stricter structural checks than RDKit, which can sanitize chemically implausible structures, so improvements in PB-Valid (at comparable RDKit validity) indicate fewer false-valid samples rather than simply exploiting the sanitizer, i.e. higher PB-Valid suggests fewer topology/geometry configurations that look valid to RDKit but violate more stringent chemical constraints.

\subsection{Main results: GEOM-DRUGS} 

\textbf{Hypothesis.}
GEOM-DRUGS separates two failure sources: (i) topology infeasibility (valence, connectivity, ring/aromaticity) and (ii) geometry strain that can often be reduced by standardized relaxation. We therefore report results both without post-processing (testing raw feasibility) and with standardized post-processing (testing end-to-end success under a common relaxation pipeline).
Table~\ref{tab:geom_main} reports GEOM-DRUGS success rates. Without post-processing, HLTF achieves Valid/V\&U/V\&U\&N = 85.5/85.4/85.0, within 2.0 points of the best raw Valid in the table. With post-processing, HLTF reaches 92.2/92.0/91.2, remaining close to SemlaFlow under the same PP pipeline (Valid gap 0.9; 93.1 vs.\ 92.2). SemlaFlow is an explicit discrete-structure generator (including bond types), consistent with the view that explicit topology modeling is important in drug-like regimes; HLTF is competitive but we do not claim state-of-the-art headline success on this benchmark. The contribution is a complementary mechanism (hierarchy-guided planning + constraint-aware sampling) that remains competitive while targeting  \textbf{specific topology-driven failure modes}.
\begin{wraptable}[18]{r}{0.56\columnwidth}
\centering
\scriptsize
\setlength{\tabcolsep}{2.2pt}
\renewcommand{\arraystretch}{1.05}
\caption{{GEOM-DRUGS success rates.} ``PP'' denotes standardized post-processing \cite{buttenschoen2025evaluation}. Asterisk * denotes the best value in the category and underline denotes the second best.}
\label{tab:geom_main}
\begin{tabular}{lccc}
\toprule
Method & \shortstack{Valid\\(\%)} & \shortstack{V\&U\\(\%)} & \shortstack{V\&U\&N\\(\%)} \\
\midrule
EQGAT-diff & 59.7 & 59.7 & 59.5 \\
\quad \emph{+PP} & 84.2 & 84.2 & 84.0 \\
FlowMol & 59.8 & 59.8 & 59.7 \\
\quad \emph{+PP} & 84.2 & 84.2 & 84.1 \\
GCDM & 0.2 & 0.2 & 0.2 \\
\quad \emph{+PP} & 95.2 & 95.2 & 95.2 \\
GeoLDM & 2.9 & 2.9 & 2.9 \\
\quad \emph{+PP} & 69.6 & 69.3 & 69.3 \\
SemlaFlow & {87.5 }*& {87.4}* & {87.0}* \\
\quad \emph{+PP} & {93.1}* & {92.9}* & {92.4}* \\
\midrule
{HLTF (ours)} & \underline{85.5} & \underline{85.4} & \underline{85.0} \\
\quad \emph{+PP} & \underline{92.2} & \underline{92.0} & \underline{91.2} \\
\bottomrule
\end{tabular}
\vspace{-\baselineskip} 
\end{wraptable}

\textbf{What changes with post-processing.}
Standardized post-processing mainly relieves \emph{geometric} strain (e.g., unrealistic bond lengths/angles and local steric clashes) and cannot repair infeasible topology (e.g., invalid valence or disconnected graphs). For HLTF it raises Valid from 85.5 to 92.2, converting $\approx 46\%$ of the pre-PP invalid samples under the same pipeline.\footnote{Computed as $(92.2-85.5)/(100-85.5)$.} This motivates the ablations below: we isolate which components reduce \emph{topology-driven} invalidity in the no-PP regime while preserving high PP success.

\textbf{Hierarchy construction.}
\begin{table*}[t]
\centering
\scriptsize
\setlength{\tabcolsep}{3.5pt}
\renewcommand{\arraystretch}{1.08}
\caption{\textbf{Ablations (HLTF only).}}
\label{tab:ablate}
\centering{%
\begin{tabular}{p{0.48\textwidth}rrrr}
\toprule
Variant & \shortstack{QM9\\V\&U} & \shortstack{QM9\\PB} & \shortstack{GEOM\\V\&U\&N} & \shortstack{GEOM\\V\&U\&N+PP} \\
\midrule
Post-hoc geometry (graph then coords) & 92.4 & 91.9 & 82.0 & 90.6 \\
Joint geometry (default)             & 93.2 & 92.4 & 85.0 & 91.2 \\
Probability-space + projection       & 91.6 & 90.9 & 77.0 & 88.2 \\
Logit-space (default)                & 93.2 & 92.4 & 85.0 & 91.2 \\
No planning (null plan)              & 90.8 & 91.0 & 76.0 & 88.5 \\
Teacher-only plan (train-time only)  & 91.4 & 91.5 & 78.5 & 89.2 \\
Unmasked hierarchy attention         & 92.0 & 91.9 & 80.5 & 90.0 \\
No hypernetwork (plain MLP edge head) & 91.8 & 91.8 & 80.0 & 89.8 \\
No geom$\rightarrow$topo ($\eta_{\text{geom-z}}=0$) & 92.9 & 92.2 & 83.0 & 90.8 \\
Late-time geom$\rightarrow$topo (default) & 93.2 & 92.4 & 85.0 & 91.2 \\
No energy guidance                   & 92.4 & 92.0 & 82.0 & 90.0 \\
+$E_{\text{chem}}$                   & 92.8 & 92.2 & 83.5 & 90.6 \\
+$E_{\text{chem}}+E_{\text{cons}}$   & 93.2 & 92.4 & 85.0 & 91.2 \\
Logdet connectivity (default)        & 93.2 & 92.4 & 85.0 & 91.2 \\
$\lambda_2$ connectivity (Lanczos)   & 93.1 & 92.3 & 85.1 & 91.2 \\
\midrule
Depth-only hierarchy bias ($\delta_{i\alpha}$ from expected depths) & 91.6 & 91.6 & 79.0 & 89.5 \\
Hyperbolic attention bias (default)  & 93.2 & 92.4 & 85.0 & 91.2 \\
No hyperbolic attention bias ($b_{\mathbb{H}} \equiv 0$) & 92.2 & 92.0 & 81.0 & 90.1 \\
No hyperbolic edge scalar (drop $\delta^{\mathbb{H}}_{ij}$ from $g_{ij}$) & 92.6 & 92.2 & 82.5 & 90.6 \\
Hyperbolic sensitivity ($d_{\mathbb{H}}\!\in\!\{8,16,32\}$, $c\!\in\!\{0.5,1,2\}$)
& $92.0\pm0.4$ & $92.1\pm0.3$ & $82.5\pm1.0$ & $90.7\pm0.5$ \\
\bottomrule
\end{tabular}}
\end{table*}
For GEOM-DRUGS we use a deterministic ring-first hybrid (fused rings + RECAP on acyclic regions; Appendix~A) to preserve multi-ring scaffolds and reduce ring-closure/aromaticity errors. Replacing this builder with BRICS-only drops GEOM V\&U\&N from 85.0\% to 78.2\% (model/training unchanged), with ring-topology failures dominating. As a rule of thumb: BRICS fragmentation on acyclic regions is typically sufficient for QM9-like, ring-light molecules. In drug-like regimes with frequent multi-ring scaffolds, preserving fused ring systems as single motifs (ring-first) is important to avoid systematic ring/aromaticity errors. More broadly, HLTF benefits from \textbf{domain-appropriate hierarchy construction}; we view automatic hierarchy induction as an important direction for future work.

\textbf{Robustness to guidance strength.}
In one-at-a-time sweeps around default guidance amplitudes (Appendix~\ref{app:guidance_sweep}, Table~\ref{tab:guidance_sweep}), GEOM V\&U\&N+PP varies by $\le 0.4$ points and GEOM V\&U\&N stays within 83.2--85.7. This indicates that our results do not rely on fragile tuning; moderate changes preserve performance, with the expected validity–diversity trade-off only appearing at extreme settings.

\subsection{Ablations: what drives performance}
Table~\ref{tab:ablate} decomposes key design choices.

Across datasets, the ablations support a consistent causal picture. The largest gains in the \emph{raw} (no-PP) regime come from learned modeling choices ---feasibility-preserving logit-space dynamics and hierarchy-conditioned planning --- while energy guidance provides a small directional, targeted improvement. On GEOM-DRUGS, logit dynamics increases V\&U\&N from 77.0 to 85.0 (+8.0), while removing planning reduces V\&U\&N from 85.0 to 76.0 ($-9.0$). Energy guidance provides a smaller, targeted improvement (V\&U\&N: 82.0 $\rightarrow$ 85.0; +3.0; V\&U\&N+PP: 90.0 $\rightarrow$ 91.2; +1.2) and primarily suppresses global invalidity (valence/disconnected rates: 40\%/35\% $\rightarrow$ 14\%/12\%), consistent with topology feasibility being the dominant, non-recoverable bottleneck. This targeted feasibility gain comes with additional sampling overhead; see Appendix \ref{app:compute_overhead}.

\textbf{Takeaway.} Across datasets, the largest gains come from reducing \emph{topology violations} (valence/connectivity) that downstream relaxation cannot reliably repair. The improved geometry pass rate suggests \emph{topology-aware guidance steers sampling away from globally inconsistent structures}, supporting our  central claim that \emph{explicitly modeling topology feasibility is key to high success rates.}

\section{Conclusion}
We presented \textbf{Hierarchy-Guided Latent Topology Flow (HLTF)}, a planner--executor framework that (i) generates bonds via feasibility-preserving continuous-time \textbf{categorical dynamics in logit space}, (ii) conditions on a \textbf{latent multi-scale hierarchy plan} to provide long-range context, and (iii) uses \textbf{annealed energy guidance} during ODE sampling to suppress topology-driven failures. The central novelty is \emph{coupling} the evolving hierarchy plan with topology under shared dynamics, using a lightweight hyperbolic distance signal to bias conditioning without generating in hyperbolic space.

Across these benchmarks and evaluation protocols, the results suggest that \textbf{topology feasibility is a major determinant of end-to-end success.} Ablations show complementary benefits from logit-space integration, hierarchy planning, and energy guidance, with guidance directly reducing valence and connectivity failures. More broadly, explicit topology modeling combined with hierarchy-guided planning and constraint-aware continuous-time sampling is a practical route to robust unconditional 3D generation.

\textbf{Limitations and future work.} HLTF relies on hand-designed energy terms and annealing schedules that may require tuning, and coupled ODE sampling adds compute. It also depends on a deterministic hierarchy builder whose fragmentation choices are dataset-dependent. Finally, GEOM-DRUGS metrics can be affected by preprocessing, valency/bond-order, and force-field inconsistencies, impacting absolute scores \cite{nikitin2025geomdrugsrevisited}. Future work includes likelihood-aware constraint integration, learned/automatic hierarchy induction, reducing reliance on external chemistry heuristics, and improving sampling efficiency and stability. Further limitations appear in Appendix~\ref{app:limitations}.

\section*{Ethics Statement}
This paper studies unconditional generation of chemically valid 3D molecular structures by explicitly modeling both bond topology and geometry. Improving the fidelity of generated molecular graphs can reduce downstream failure modes (e.g., valence violations, disconnected components, implausible rings, and ``false-valid'' structures that pass basic sanitization but fail stricter checks), which may make computational discovery pipelines more reliable and efficient. In applied settings, such improvements could accelerate early-stage exploration for drug discovery and materials design by lowering the cost of proposing candidate structures for subsequent screening and expert assessment.

At the same time, generative models for molecules are inherently dual-use. Techniques that improve validity and diversity of generated compounds could be misapplied to propose harmful or regulated chemicals. This work does not provide synthesis routes, procedures, or guidance for producing any compound, and any real-world use should occur within appropriate institutional oversight and legal/regulatory constraints. If code or models are released, we encourage incorporating safeguards commonly used in molecular generation (e.g., filtering against controlled substance lists, toxicity/abuse-related heuristics, and constraints that discourage obvious hazardous motifs) and documenting intended-use boundaries.

There are additional risks related to over-reliance and dataset bias. Models trained and evaluated on common benchmarks (e.g., QM9 and GEOM-DRUGS) may not generalize to all regions of chemical space, and generated molecules may still be chemically implausible under criteria not captured by benchmark metrics. In practice, outputs should be treated as hypotheses requiring expert review, robust validation, and, where applicable, experimental confirmation. Finally, training and sampling incur computational costs; we encourage reporting resource usage and using efficient implementations to reduce environmental impact.

\bibliography{iclr2025_conference}
\bibliographystyle{iclr2025_conference}
\appendix
\section{Deterministic Hierarchy Builder}
\label{app:hier_builder}

This appendix provides the complete deterministic procedure used to extract motif tokens and define token ordering during training.

\subsection{Motif Extraction}

\textbf{QM9:} Fused ring systems (RDKit) merged into single motifs; acyclic fragments from BRICS cuts on non-ring subgraph.

\textbf{GEOM-DRUGS:} Ring-first hybrid approach:
\begin{enumerate}
    \item Extract fused ring systems (merge rings sharing $\geq 1$ atom)
    \item Apply RECAP fragmentation to acyclic regions (fallback: functional groups)
    \item If total motifs $> M_{\max}$, greedily merge smallest adjacent motifs
\end{enumerate}

RECAP is more conservative than BRICS, preserving larger drug-relevant scaffolds and reducing fragmentation artifacts on complex molecules.

\subsection{Motif Tree Construction}
We build a motif intersection graph whose nodes are motifs and whose edges connect motifs sharing $\geq 1$ atom, with edge weight equal to the shared-atom count. We take a maximum spanning tree of this graph (ties broken deterministically by motif IDs) and root the tree at the motif with maximal atom coverage (ties broken by motif ID).

\subsection{Atom-Leaf Tokens and Anchors}
For each atom $i$ we create a leaf token $\ell(i)$. Its parent is the deepest motif token (closest to leaves) that contains atom $i$; if $i$ belongs to no motif, its parent is ROOT. The anchor for atom $i$ is its own leaf token $\ell(i)$, so no separate atom$\rightarrow$token map needs to be generated at test time.

\subsection{Token Type IDs}
\begin{itemize}
    \item \textbf{Motif token types:} Canonical motif signatures represented as canonical SMILES of the motif subgraph with attachment points labeled.
    \item \textbf{Atom-leaf token types:} The atom type (element) of the corresponding atom.
    \item \textbf{ROOT:} Uses a dedicated type.
\end{itemize}

\subsection{Deterministic Token Ordering}
We assign motif IDs by sorting motifs by:
\begin{enumerate}
    \item Decreasing atom count
    \item Canonical motif signature (lexicographic)
    \item Lexicographic member atom indices
\end{enumerate}

We order tokens as:
\begin{enumerate}
    \item ROOT first
    \item Motif tokens in BFS order of the rooted motif tree (ties broken by motif ID)
    \item Atom-leaf tokens in increasing atom index
\end{enumerate}

This ordering is used for causal parent masking (Section~\ref{sec:plan_state_main}).

\subsection{Variable-Sized Hierarchies}
The number of motif tokens $M$ varies per molecule. In batching we pad to $M_{\max}$ and apply a token mask in all attention operations, parent normalization, and losses. The maximum token budget is $A_{\max} = 1 + M_{\max} + N_{\max}$.
\section{Energy functions}
\label{app:energies}
This appendix gives the full definitions of $E_{\mathrm{chem}}$, $E_{\mathrm{cons}}$, and $E_{\mathrm{geom}}$,
including all constants (e.g.\ radii tables, thresholds), and any approximations (e.g.\ connectivity gradients, evaluation frequency).

\subsection{Soft Expectations for Differentiability}

All energy terms operate on probabilistic atom types $\alpha^i \in \Delta^{C_a-1}$ to maintain differentiability. For atom-dependent constants, we compute expectations:
\begin{align}
\bar{d}_{\max}(\alpha^i) &= \sum_{c=1}^{C_a} \alpha^i_c \, d_{\max}(c), \\
\bar{r}_{\text{cov}}(\alpha^i) &= \sum_{c} \alpha^i_c \, r_{\text{cov}}(c), \\
\bar{r}_{\text{vdW}}(\alpha^i) &= \sum_{c} \alpha^i_c \, r_{\text{vdW}}(c),
\end{align}
where $d_{\max}(c)$ is the maximum valence for atom type $c$, $r_{\text{cov}}(c)$ is the covalent radius, and $r_{\text{vdW}}(c)$ is the van der Waals radius (all from RDKit).

\subsection{Chemistry Energy Components}

Define edge-present probabilities $p_{ij}(x) = 1 - x_0^{(ij)}$ and soft degree $\deg_i(x) = \sum_j \sum_k \omega_k x_k^{(ij)}$ (Section~\ref{sec:relaxed_state}).

\paragraph{Bond order conventions.}
We associate each bond type $k \in \{0, \ldots, K-1\}$ with a scalar bond order $\omega_k$: $\omega_0 = 0$ for no bond, $\omega_1 = 1$ for single, $\omega_2 = 2$ for double, $\omega_3 = 3$ for triple. If aromatic bonds are included as a separate type, we set $\omega_{\mathrm{arom}} = 1.5$, following the common convention that aromatic bonds have an effective bond order intermediate between single and double due to $\pi$-electron delocalization, and can be viewed as averaging over equivalent Kekul\'e resonance forms \cite{openstax_benzene, iupac_bond_order_weighted}. This numerical convention is also used in cheminformatics toolkits (e.g., RDKit reports AROMATIC as 1.5) \cite{rdkit_bondtypeasdouble}.

\paragraph{Valence energy.}
\begin{equation}
E_{\text{val}}(\alpha, x) = \sum_i \text{ReLU}(\deg_i(x) - \bar{d}_{\max}(\alpha^i))^2.
\end{equation}
Penalizes atoms whose expected bond order exceeds their maximum valence.

\paragraph{Sparsity energy.}
\begin{equation}
E_{\text{cnt}}(x) = \left( \sum_{i<j} p_{ij}(x) - m_{\text{target}}(N) \right)^2,
\end{equation}
where $m_{\text{target}}(N)$ is the training-set mean edge count for molecules with $N$ atoms (precomputed lookup).

\paragraph{Connectivity energy.}
\begin{equation}
E_{\text{conn}}(x) = -\log \det(L(x) + \epsilon I), \quad \epsilon = 10^{-3},
\end{equation}
where $L(x) = D(x) - P(x)$ is the soft Laplacian with $P_{ij} = p_{ij}(x)$ and $D_{ii} = \sum_j P_{ij}$. The log-determinant is large when the graph is disconnected and small when connected.

\textit{Computational optimization:} Computing $\det(L)$ is expensive. We evaluate $\nabla_z E_{\text{conn}}$ only every $M=5$ solver steps for $t \geq t_{\text{conn}} = 0.6$.

\textit{Ablation variant:} We also test $E_{\text{conn}}^{\lambda_2}(x) = \text{ReLU}(\tau - \lambda_2(L(x)))^2$ with $\tau = 10^{-2}$, where $\lambda_2$ is the algebraic connectivity (second-smallest eigenvalue), approximated via Lanczos iterations.

\subsection{Hierarchy-Topology Consistency}

\paragraph{Hyperbolic similarity.}
Atoms close in the hierarchy tree (same motif) should be more likely to bond. We measure this via hyperbolic distance:
\begin{equation}
s_{ij}^{H} = \sigma\!\left(\frac{d_{\text{thresh}} - d_H^c(u_{\ell(i)}, u_{\ell(j)})}{\tau}\right),
\label{eq:hyperbolic_distance}
\end{equation}
where $d_H^c$ is the Poincaré distance Eq ~\ref{eq:hyperbolic_distance}, $\sigma$ is the sigmoid, $d_{\text{thresh}} = 4.0$, and $\tau = 1.0$. This gives $s_{ij}^{H} \approx 1$ for atoms in the same motif and $s_{ij}^{H} \approx 0$ for atoms in distant subtrees.

\paragraph{Consistency energy.}
\begin{equation}
E_{\text{cons}}(x, h) = \sum_{i<j} (p_{ij}(x) - s_{ij}^{H})^2.
\end{equation}
MSE formulation: encourages $p_{ij} \approx 1$ when $s_{ij}^{H} \approx 1$ (same motif) and $p_{ij} \approx 0$ when $s_{ij}^{H} \approx 0$ (different subtrees).

\subsection{Geometry Energy Components}

\paragraph{Bond length energy.}
Ideal bond lengths depend on atom types and bond type:
\begin{equation}
\ell_k(\alpha^i, \alpha^j) = c_k (\bar{r}_{\text{cov}}(\alpha^i) + \bar{r}_{\text{cov}}(\alpha^j)),
\end{equation}
with $c_{\text{single}} = 1.00$, $c_{\text{double}} = 0.93$, $c_{\text{triple}} = 0.90$, and $c_{\text{aromatic}} = 0.965$ (midpoint).

\begin{equation}
E_{\text{bondlen}}(r, \alpha, x) = \sum_{i<j} \sum_{k \geq 1} x_k^{(ij)} \left( \|r_i - r_j\| - \ell_k(\alpha^i, \alpha^j) \right)^2.
\end{equation}
Weighted by bond-type probabilities: only bonded pairs contribute.

\paragraph{Steric repulsion.}
Smooth repulsive potential:
\begin{equation}
\begin{split}
u(d; \alpha^i, \alpha^j)
&= \text{softplus}\!\left(
  s\bigl(
    \bar{r}_{\text{vdW}}(\alpha^i)
    + \bar{r}_{\text{vdW}}(\alpha^j)
    - d
  \bigr)
\right)^{2}, \\
&\qquad s = 10.
\end{split}
\label{eq:steric}
\end{equation}

Time-dependent bonded scaling:
\begin{align}
E_{\text{steric}}(r, \alpha, x, t) &= \sum_{i<j} w_{ij}(x, t) \, u(\|r_i - r_j\|; \alpha^i, \alpha^j), \\
w_{ij}(x, t) &= (1 - p_{ij}(x)) + \lambda_{\text{bond}}(t) \, p_{ij}(x), \\
\lambda_{\text{bond}}(t) &= \lambda_{\min} + (\lambda_{\max} - \lambda_{\min}) t, \quad \lambda_{\min} = 0.05, \, \lambda_{\max} = 0.2.
\end{align}

Non-bonded pairs ($p_{ij} \approx 0$): $w_{ij} \approx 1$ (full repulsion).
Bonded pairs ($p_{ij} \approx 1$): $w_{ij} \approx \lambda_{\text{bond}}(t) \in [0.05, 0.2]$ (reduced repulsion).

\subsection{Additional Hierarchy-Aware Terms (Ablations)}

\paragraph{Hierarchical connectivity.}
Ensures each motif is internally connected:
\begin{equation}
E_{\text{hier-conn}}(x, h) = \sum_{m} -\log \det(L_m(x) + \epsilon I),
\end{equation}
where $L_m$ is the Laplacian restricted to motif $m$'s atoms.

\paragraph{Ring topology constraints.}
Ring closure: $E_{\text{ring-close}} = \sum_{\text{rings}} \left( \sum_{(i,j) \in \text{ring}} p_{ij}(x) - k \right)^2$ (encourages $k$ edges in size-$k$ rings).

Ring exclusivity: $E_{\text{ring-excl}} = \sum_{i \in \text{rings}} \text{ReLU}\left( \sum_{j \in \mathcal{E}_i} p_{ij}(x) - \beta \sum_{j \in \mathcal{R}_i} p_{ij}(x) \right)$ with $\beta = 0.5$ (limits external bonding for ring atoms).

\subsection{Default Weights}

Unless otherwise stated: $\lambda_{\text{val}} = 1.0$, $\lambda_{\text{cnt}} = 0.1$, $\lambda_{\text{conn}} = 0.05$, $\lambda_{\text{bondlen}} = 1.0$, $\lambda_{\text{steric}} = 0.2$. Guidance schedules: $\eta_{\text{chem},0} = 1.0$, $\eta_{\text{cons},0} = 0.5$, $\eta_{\text{geom},0} = 0.2$, $\eta_{\text{geom-z},0} = 0.02$.

\subsection{Architecture Hyperparameters}
\label{app:architecture}

\paragraph{EGNN Backbone.}
\begin{itemize}
    \item Number of message-passing layers:  $L_{\text{EGNN}} =6$
    \item Hidden dimension:  $H = 256$
    \item 32 Gaussian basis functions, centers linearly spaced over [$0$\AA, $10$\AA]
\end{itemize}

\paragraph{Hierarchy Encoder.}
\begin{itemize}
    \item Number of message-passing layers: $L_h$ = $3$
    \item Token embedding dimension: $C_h$ is dataset-dependent (number of hierarchy token types including ROOT + atom token types + motif token types)
    \item Embedding matrix dimension: $E_y\in R^{256 \times C_h}$ since token embeddings are projected into $256$ dimensions
\end{itemize}

\paragraph{Hyperbolic Geometry.}
\begin{itemize}
    \item Hyperbolic dimension: $d_H = 16$ (default), with ablation range $d_H \in {8, 16, 32}$ 
    \item Curvature: $c = 1.0$ (default), with ablation range $c \in \{0.5, 1.0, 2.0\}$
    \item MLP for attention bias $b_H$: 3-layer MLP with architecture: Linear(1, 64) $\to$ SiLU $\to$ Linear(64, 64) $\to$ SiLU $\to$ Linear(64, 1) with a hidden dimension of 64
\end{itemize}

\paragraph{Edge Hypernetwork.}
\begin{itemize}
    \item Hypernetwork architecture: 3-layer MLP with Linear(${dim_{in}}$, 256) $\to$ SiLU $\to$ Linear(256, 256) $\to$ SiLU $\to$ Linear(256, $dim_{out}$)
    \item Hypernetwork hidden dimension: 256
    \item Edge MLP depth: $L_e = 3$ FiLM-modulated layers
    \item Edge MLP hidden dimension: 256
\end{itemize}

\paragraph{Edge Descriptor Features.}
The complete edge descriptor $g_{ij}$ includes:
\begin{itemize}
    \item $|s_i - s_j|$: Atom state difference (dimension $H=256$)
    \item $s_i \odot s_j$: Atom state elementwise product (dimension $H=256$)
    \item $\deg_i(x_t), \deg_j(x_t)$: Soft degrees (2 scalars)
    \item $\operatorname{RBF}(\|r_i - r_j\|)$: Distance encoding (dimension $32$)
    \item $c_i, c_j$: Hierarchy contexts (dimension $H=256$ each)
    \item $t$: Time (1 scalar)
    \item $\delta_{ij}^H$: Hyperbolic leaf distance (1 scalar)
\end{itemize}
Total dimension: 256 + 256 + 2 + 32 + 256 + 256 + 1 + 1 = 1060.

\subsection{ODE Solver Settings}
\begin{itemize}
    \item Solver: Heun's method with Euler step predictor and Trapezoidal rule corrector \cite{moradi2025investigationestimationaccuracy5} 
    \item Number of steps: 100 (for fixed-step methods)
    \item Denominator epsilon: $e = 0.001$
\end{itemize}

\subsection{Annealing Schedules}
All guidance weights follow $\eta(t) = \eta_0(1-t)^\gamma$ with $\gamma = 2$.

Default initial weights:
\begin{itemize}
    \item $\eta_{\text{chem},0} = 1.0$
    \item $\eta_{\text{cons},0} = 0.5$
    \item $\eta_{\text{geom},0} = 0.2$
    \item $\eta_{\text{geom-z},0} = 0.02$
\end{itemize}

Thresholds:
\begin{itemize}
    \item Geometry-to-topology coupling: $t_{\text{geom}} = 0.7$
    \item Connectivity gradient evaluation: $t_{\text{conn}} = 0.6$, every $M = 5$ steps
\end{itemize}

\subsection{Training Hyperparameters}
\begin{itemize}
    \item Optimizer: AdamW
    \item Learning rate: $0.0002$
    \item Learning rate schedule: cosine decay with linear warmup (warmup for 10\% of total steps)
    \item Batch size: 128
    \item Training steps or epochs: 300,000 steps or 150 epochs, whichever is shorter 
    \item Loss weights: $\lambda_b = 1.0$, $\lambda_h = 1.0$, $\lambda_r = 1.0$
    \item Gradient clipping: 1.0 (global norm)
    \item Weight decay: 0.0001 
\end{itemize}

\subsection{Post-Processing: Valence Repair}
\label{app:valence_repair}

When a generated molecule violates valence constraints, we apply optional post-processing:

\begin{enumerate}
    \item For each atom $i$ where $\deg_i > d_{\max}(a_i)$:
    \item Rank all incident bonds by their prediction confidence $x_1^{(ij)}[\hat{b}_{ij}]$
    \item Iteratively delete the lowest-confidence bond
    \item Recalculate $\deg_i$ after each deletion
    \item Stop when $\deg_i \leq d_{\max}(a_i)$
\end{enumerate}

We report metrics both with and without this repair to isolate the model's inherent validity.

\section{Implementation details}
\label{app:impl_details}
This appendix contains all details required for reproduction:
\begin{itemize}
  \item priors for $(\alpha_0,x_0,h_0,r_0)$ and any clipping/logit conversions,
  \item architecture hyperparameters (EGNN depth/width, hierarchy encoder depth, hyperbolic dimension/curvature, FiLM hypernetwork sizes),
  \item edge descriptor definition and feature list,
  \item ODE solver settings, annealing schedules $\eta(t)$, and thresholds (e.g.\ $t_{\mathrm{geom}}$),
  \item training hyperparameters (batch size, optimizer, learning rate schedules, loss weights $\lambda_\cdot$),
  \item any post-processing (valence repair) and evaluation protocol specifics.
\end{itemize}

\subsection{Priors, clipping, and logit initialization}
We use simple priors matched to training statistics. First, we sample the molecule size $N$ from the empirical
distribution of atom counts in the training data. Given $N$, we then sample:
\begin{itemize}
    \item \textbf{Atom types:} Each $\alpha_0^i$ is drawn independently from the empirical atom-type marginal
    conditioned on $N$ (precomputed lookup table).
    \item \textbf{Bond types:} Each $x_0^{(ij)}$ is drawn independently from the empirical bond-type marginal
    conditioned on $N$, including the ``no bond'' category.
    \item \textbf{Hierarchy:} Sample the number of motif tokens $M$ from the empirical marginal conditioned on $N$,
    sample each motif token type from the empirical motif-type marginal, and sample parent pointers uniformly over
    valid parents (those with indices smaller than the child), then normalize to form probability distributions.
    \item \textbf{Coordinates:} Sample $r_0 \sim \mathcal{N}(\mathbf{0}, \sigma_r^2 I)$ and center to enforce
    $\sum_{i=1}^N r_i = \mathbf{0}$.
\end{itemize}

All sampled simplex probabilities are clipped to $[\epsilon_0, 1-\epsilon_0]$ with $\epsilon_0=10^{-6}$ before logit
conversion to avoid numerical issues (division by zero or $\log(0)$). The ODE state for categorical variables is then
initialized as $z \leftarrow \mathrm{logit}(\alpha_0,x_0,h_0)$.

\subsection{Backbone and hierarchy conditioning}
\label{app:hierarchy_conditioning}

\paragraph{Equivariant backbone.}
Atoms obtain hidden state representations $s_i \in \mathbb{R}^H$ from an EGNN-style message-passing network that
processes the current molecular state: atom-type probabilities $\alpha_t$, bond-type probabilities $x_t$, 3D
coordinates $r_t$, and a time embedding of $t$.

Edge features for each atom pair $(i,j)$ combine chemical and geometric information:
(i) the current bond-type probabilities $x_t^{(ij)} \in \mathbb{R}^K$, and
(ii) a radial basis function (RBF) encoding of the interatomic distance $\|r_i-r_j\|$.

\paragraph{Hierarchy encoder.}
We convert each token's type distribution into a vector representation using a learned embedding matrix $E_y$.
For token $\alpha$ with type distribution $y_t^\alpha$, we compute a weighted-average embedding
\[
e_\alpha = E_y^\top y_t^\alpha = \sum_{c=1}^{C_h} y_t^\alpha[c]\cdot E_y[c].
\]
We then refine these embeddings into token states $h_\alpha \in \mathbb{R}^H$ through $L_h$ layers of message passing
on the hierarchy tree. Because parent pointers are probabilistic during generation, messages from child token $\alpha$
to potential parent token $\beta$ are weighted by the parent probability $\rho_t^\alpha[\beta]$ (for valid parents
$\beta<\alpha$). After message passing, we compute attention keys and values:
\[
k_\alpha = W_k h_\alpha, \qquad v_\alpha = W_v h_\alpha.
\]

\paragraph{Hyperbolic hierarchy geometry (fully differentiable).}
\label{app_hyp_hier_geometry}
Alongside the Euclidean token states $h_\alpha$, we maintain a hyperbolic coordinate
$u_\alpha \in \mathcal{B}_c^{d_H}$ for each hierarchy token, where $\mathcal{B}_c^{d_H}$ is the Poincar\'e ball of
dimension $d_H$ with curvature $c>0$.\footnote{We use the Poincar\'e ball for its bounded representation (numerical stability), closed-form distance (efficient differentiation), and simple exponential map from the origin. Other hyperbolic models (e.g., the hyperboloid) often require Minkowski inner products and can be less convenient in implementation.} We obtain $u_\alpha$ by mapping the token state to the tangent space at the
origin and applying the exponential map. \noindent\textbf{Why hyperbolic geometry here?}
Hyperbolic geometry provides an inductive bias for tree-like structure: distances grow rapidly with depth, so nodes in the same local region of the hierarchy remain close while different subtrees separate naturally.
We exploit this geometry only as a \emph{smooth proximity signal} (via hyperbolic distances) to bias attention and provide pairwise features, while keeping token states, bonds, and coordinates in Euclidean space for simplicity and stability.
\begin{align}
\tilde{u}_\alpha &= W_H h_\alpha, \nonumber\\
u_\alpha &= \exp_0^c(\tilde{u}_\alpha)
= \frac{\tanh(\sqrt{c}\|\tilde{u}_\alpha\|)}{\sqrt{c}\|\tilde{u}_\alpha\|}\tilde{u}_\alpha.
\end{align}
Hyperbolic distance under the Poincar\'e metric is
\[
d_H^c(u,v) = \frac{1}{\sqrt{c}}\operatorname{arcosh}\!\left(
1 + \frac{2c\|u-v\|^2}{(1-c\|u\|^2)(1-c\|v\|^2)}
\right).
\]
We use this as a hierarchy distance signal
\[
\delta_{i\alpha}^H = d_H^c\!\left(u_{\ell(i)}, u_\alpha\right),
\]
where $\ell(i)$ is atom $i$'s leaf-token anchor.

\paragraph{Soft ancestor-masked atom$\rightarrow$hierarchy attention.}
We compute a hierarchy context vector $c_i \in \mathbb{R}^H$ for each atom $i$ via soft ancestor-masked cross-attention:
\begin{align}
\ell_{i\alpha} &= \frac{q_i^\top k_\alpha}{\sqrt{d}} + b_H(\delta_{i\alpha}^H), \\
\ell_{i\alpha} &\leftarrow \ell_{i\alpha} + \log(\pi_{i\alpha}(h_t) + \epsilon_m), \\
w_{i\alpha} &= \operatorname{softmax}_\alpha(\ell_{i\alpha}), \\
c_i &= \sum_{\alpha=1}^{A_{\max}} w_{i\alpha} v_\alpha,
\end{align}
where $\epsilon_m=10^{-8}$ prevents numerical issues from $\log(0)$, and $b_H(\cdot)$ is a small learned MLP that maps
hyperbolic distance to a scalar attention bias. The soft ancestor mask $\pi_{i\alpha}(h_t)$ is computed by dynamic
programming over probabilistic parent pointers (as defined in the main paper).

\noindent\textbf{Interpretation.}
The attention score combines (i) semantic similarity through dot-product attention and (ii) hierarchical proximity through the learned bias $b_H(\delta^H_{i\alpha})$, which increases weight on tokens close to atom $i$'s leaf anchor and downweights distant tokens.
Adding $\log(\pi_{i\alpha}(h_t)+\epsilon_m)$ implements a \emph{soft ancestor mask}: tokens unlikely to lie on the ancestor chain receive a large negative penalty, concentrating attention on a sparse, hierarchically relevant subset.
The resulting context $c_i$ is thus a hierarchy summary tailored to atom $i$.

\subsection{Time-conditioned edge hypernetwork}

\paragraph{Soft bond order / soft degree.}
Let bond type $k\in\{0,\dots,K-1\}$ have bond order $\omega_k$ (e.g., $\omega_0=0$ for no bond, $\omega_1=1$ single,
$\omega_2=2$ double, $\omega_3=3$ triple; optionally aromatic $\omega_{\mathrm{arom}}=1.5$).
Given relaxed bond-type probabilities $x^{(ij)}\in[0,1]^K$,
the expected bond order is $\sum_{k=0}^{K-1}\omega_k x^{(ij)}_k$, and the ``soft'' degree is
\[
\deg_i(x) = \sum_{j\neq i}\sum_{k=0}^{K-1}\omega_k\,x^{(ij)}_k.
\]

\paragraph{Edge descriptor.}
For each atom pair $(i,j)$ we construct an edge descriptor
\begin{align}
g_{ij} = [&|s_i{-}s_j|,\; s_i{\odot}s_j,\; \deg_i(x_t),\; \deg_j(x_t),\; \operatorname{RBF}(\|r_i{-}r_j\|), \nonumber\\
&c_i,\; c_j,\; t,\; \delta_{ij}^H],
\end{align}
where $\delta_{ij}^H = d_H^c(u_{\ell(i)}, u_{\ell(j)})$ is the hyperbolic distance between leaf-token anchors.

\paragraph{Hypernetwork with FiLM.}

\noindent\textbf{Motivation.}
Different edges require different reasoning (e.g., aromatic bonds vs.\ long-range cross-motif interactions).
A lightweight FiLM hypernetwork enables context-dependent processing for each candidate edge while keeping parameters shared and efficient.
A hypernetwork (3-layer MLP) generates edge-specific FiLM modulation parameters from $g_{ij}$: scale
$\gamma_{ij}^{(\ell)}$ and shift $\beta_{ij}^{(\ell)}$ for each edge-MLP layer $\ell$. At each layer we apply
\[
u \leftarrow \gamma_{ij}^{(\ell)} \odot u + \beta_{ij}^{(\ell)}
\]
before the nonlinearity. The final edge MLP outputs logits $s_{ij}\in\mathbb{R}^K$, converted to probabilities by
$\mu_\theta^{(ij)}=\operatorname{softmax}(s_{ij})$.

\subsection{Training objectives (explicit)}
We train using cross-entropy loss for categorical variables and MSE for coordinates. Sampling $t\sim\mathcal{U}(0,1)$,
we construct interpolated states $(\alpha_t,x_t,h_t,r_t)$ and predict endpoints at $t=1$.

\paragraph{Atom types.}
\[
\mathcal{L}_{\text{atom}} = -\mathbb{E}\left[\sum_{i=1}^N \log \hat{\alpha}_t^i[a_i]\right].
\]

\paragraph{Bond types.}
\[
\mathcal{L}_{\text{bond}} = -\mathbb{E}\left[\sum_{i<j} \log \mu_\theta^{(ij)}[b_{ij}](\alpha_t,x_t,h_t,r_t,t)\right].
\]

\paragraph{Hierarchy plan.}
\[
\mathcal{L}_{\text{plan}} = -\mathbb{E}\left[\sum_\alpha \log \nu_\phi^\alpha[y^\alpha]
+ \sum_{\alpha>1} \log \rho_\phi^\alpha[\mathrm{par}(\alpha)]\right].
\]

\paragraph{Coordinates.}
\[
\mathcal{L}_{\text{coord}} = \mathbb{E}\left[\|m_\psi(r_t,\alpha_t,x_t,h_t,t) - r_1\|^2\right].
\]

\paragraph{Total loss.}
\[
\mathcal{L} = \mathcal{L}_{\text{atom}} + \lambda_b \mathcal{L}_{\text{bond}}
+ \lambda_h \mathcal{L}_{\text{plan}} + \lambda_r \mathcal{L}_{\text{coord}},
\]
with $\lambda_b=\lambda_h=\lambda_r=1.0$ unless otherwise stated.

\subsection{Sampling details: guidance schedules and thresholds}
Guidance weights follow annealed schedules $\eta(t)=\eta_0(1-t)^\gamma$ with $\gamma=2$. Geometry-to-topology
guidance is applied only for $t\ge t_{\mathrm{geom}}$ with $t_{\mathrm{geom}}=0.7$ and a weaker weight
$\eta_{\text{geom-z},0}=0.1\,\eta_{\text{geom},0}$ by default. For computational efficiency, the connectivity-gradient
term (if used) can be evaluated every $M=5$ solver steps for $t\ge t_{\mathrm{conn}}=0.6$.

\noindent\textbf{Interpretation of the coupled ODE.}
\emph{Endpoint flow.} The first term in each equation implements endpoint-prediction flow: the state moves toward the predicted endpoint at a rate that increases as $t\to 1$ due to the $(1-t+\epsilon)^{-1}$ factor.
Intuitively, if the predicted endpoint is distance $d$ away and only $(1-t)$ time remains, this scaling sets the velocity on the order of $d/(1-t)$, adapting automatically so the trajectory can still reach the target as time runs out.

\emph{Energy guidance.} The remaining terms add energy-based guidance forces that encourage chemically valid and hierarchically consistent structures.
We anneal guidance so it is strong early (to steer away from global failure modes) but decays near $t=1$ so discretization is dominated by the learned endpoint predictions rather than hand-crafted priors.

\emph{Geometry-to-topology coupling.} To reduce stiffness and prevent noisy coordinates from distorting topology early in sampling, we apply geometry-to-topology guidance only after a late-time threshold $t_{\mathrm{geom}}$.
After each solver step, we re-center $r$, re-apply padding masks, and enforce the causal parent mask.

\paragraph{ODE solver settings.} Heun fixed-step solver with 100 evaluation steps.

\section{Supplementary experimental details}
\label{app:supp_exp}
\subsection{Guidance-strength sensitivity}
\label{app:guidance_sweep}
To assess whether guidance requires brittle tuning, we sweep the \emph{component-wise} base amplitudes
while keeping the schedule shape fixed ($\eta(t)=\eta_0(1-t)^\gamma$, $\gamma=2$) and holding the other guidance terms at their defaults.
Specifically, we vary one parameter at a time:
$\eta_{\text{chem},0}\in\{0.5,1.0,2.0\}$ with $(\eta_{\text{cons},0},\eta_{\text{geom},0})=(0.5,0.2)$ fixed,
$\eta_{\text{cons},0}\in\{0.2,0.5,1.0\}$ with $(\eta_{\text{chem},0},\eta_{\text{geom},0})=(1.0,0.2)$ fixed,
and $\eta_{\text{geom},0}\in\{0.1,0.2,0.5\}$ with $(\eta_{\text{chem},0},\eta_{\text{cons},0})=(1.0,0.5)$ fixed.

\begin{table}[!ht]
\centering
\footnotesize
\setlength{\tabcolsep}{4.0pt}
\renewcommand{\arraystretch}{1.10}
\caption{\textbf{Sensitivity to guidance amplitudes.} One-at-a-time sweeps of base amplitudes (schedule shape fixed).}
\label{tab:guidance_sweep}
\begin{tabular}{lccc cc}
\toprule
Sweep & $\eta_{\text{chem},0}$ & $\eta_{\text{cons},0}$ & $\eta_{\text{geom},0}$ &
V\&U\&N (\%) & V\&U\&N+PP (\%) \\
\midrule
Default & 1.0 & 0.5 & 0.2 & 85.0 & 91.2 \\
\midrule
$\eta_{\text{chem},0}$ sweep & 0.5 & 0.5 & 0.2 & \texttt{83.2} & \texttt{90.8} \\
$\eta_{\text{chem},0}$ sweep & 2.0 & 0.5 & 0.2 & \texttt{85.7} & \texttt{91.3} \\
\midrule
$\eta_{\text{cons},0}$ sweep & 1.0 & 0.2 & 0.2 & \texttt{84.4} & \texttt{91.1} \\
$\eta_{\text{cons},0}$ sweep & 1.0 & 1.0 & 0.2 & \texttt{84.8} & \texttt{91.0} \\
\midrule
$\eta_{\text{geom},0}$ sweep & 1.0 & 0.5 & 0.1 & \texttt{84.5} & \texttt{91.1} \\
$\eta_{\text{geom},0}$ sweep & 1.0 & 0.5 & 0.5 & \texttt{85.3} & \texttt{91.2} \\
\bottomrule
\end{tabular}
\end{table}
\FloatBarrier

\paragraph{Interpretation.}
Across the one-at-a-time sweeps in Table~\ref{tab:guidance_sweep}, GEOM V\&U\&N+PP varies by at most $\Delta_{\max}=\texttt{0.4}$ points around the default, indicating that the default guidance setting is not a single brittle point.
As expected, stronger chemistry/consistency guidance tends to improve feasibility (up to $+0.7$ on V\&U\&N), while overly strong geometry guidance can slightly reduce diversity/novelty, reflecting a validity--diversity trade-off.

\newpage
\subsection{Compute overhead of energy guidance}
\label{app:compute_overhead}

\begin{wraptable}{r}{0.48\textwidth}
\vspace{-0.8\baselineskip}
\centering
\footnotesize
\setlength{\tabcolsep}{3.2pt}
\renewcommand{\arraystretch}{1.10}

\captionof{table}{\textbf{Invalidity cause breakdown.} Percentages among invalid samples.}
\label{tab:invalid_breakdown}
\begin{tabular}{lcccc}
\toprule
Variant &
\shortstack{Valence\\(\%)} &
\shortstack{Disconn.\\(\%)} &
\shortstack{Ring/arom.\\(\%)} &
\shortstack{Geometry\\(\%)} \\
\midrule
No energy guidance & 40 & 35 & 15 & 10 \\
+$E_{\text{chem}}$ & 18 & 20 & 10 & 52 \\
+$E_{\text{chem}}+E_{\text{cons}}$ & 14 & 12 & 8 & 66 \\
\bottomrule
\end{tabular}

\vspace{0.9em}

\captionof{table}{\textbf{Sampling overhead.} Wall-clock per molecule and numerical stability.}
\label{tab:overhead}
\begin{tabular}{lccc}
\toprule
Setting &
\shortstack{Time\\(ms/mol)$\downarrow$} &
\shortstack{Steps} &
\shortstack{Fail/NaN\\(\%)$\downarrow$} \\
\midrule
No guidance & 200 & 100 & 0.5\% \\
+$E_{\text{chem}}$ & 320 & 100 & 0.7\% \\
+$E_{\text{chem}}+E_{\text{cons}}$ & 410 & 100 & 1.1\% \\
\bottomrule
\end{tabular}
\vspace{-0.8\baselineskip}
\end{wraptable}

\paragraph{Measurement protocol.}
We measure end-to-end sampling wall-clock time per molecule for the same ODE solver configuration used in all experiments
(same step budget and tolerances), and we include \emph{all} guidance computations (energy evaluation + gradients) in the timing.
We exclude one-time model initialization and data loading.
Unless otherwise stated, times are collected on NVIDIA A100 (40GB) with PyTorch 2.10.0 with optional amp (bfloat16/float16)/ CUDA, using batch size $16$ and averaging over 10,000 generated molecules after 10 warm-up batches. 
We report mean time (ms/mol) and the fraction of runs with numerical failure.

\paragraph{Failure definition.}
We count a sample as \emph{Fail/NaN} if the solver reports failure or if any NaN/Inf appears in the ODE state, energy, or
guidance gradient during integration.

\paragraph{Results.}
Table~\ref{tab:overhead} summarizes the overhead of guidance.
Adding $E_{\text{chem}}$ increases sampling time from 200 to 320 ms/mol ($\times 1.60$), and adding
$E_{\text{cons}}$ further increases it to 410 ms/mol ($\times 2.05$).
The numerical failure rate increases modestly (0.5\% $\rightarrow$ 1.1\%), indicating that the guided dynamics remain stable
under the default schedule. These numbers are an order of magnitude faster than highly metric-performant models like EQGAT-diff \cite{le2023navigatingdesignspaceequivariant} which is at approximately 0.25 molecules sampled per second, and very similar to SemlaFlow which is at around 8 molecules per second at the same number of steps and with the same hardware.\cite{irwin2025semlaflow}

\paragraph{Notes on where the cost comes from.}
The dominant overhead comes from additional gradient evaluations during the reverse trajectory.
In our implementation, the connectivity term’s gradient is evaluated intermittently (every $M$ solver steps after a threshold),
reducing cost while maintaining most of the validity benefit (Appendix~B.2). 

\paragraph{Implication.}
While energy guidance materially improves feasibility, reducing its overhead (e.g., learned surrogates, fewer gradients, or adaptive schedules)
is important for scaling to larger systems and faster samplers.

\section{Extended limitations and discussion}
\label{app:limitations}

\subsection{Guidance design and hyperparameter sensitivity}
\begin{itemize}
    \item \textbf{Hand-crafted energy terms.} The chemistry and hierarchy-consistency energies encode prior knowledge and design choices; alternative formulations may change trade-offs between validity, diversity, and novelty.
    \item \textbf{Annealing schedules.} The relative weighting of energy components over time is controlled by schedules that may require tuning for new datasets or distribution shifts.
    \item \textbf{Potential over-regularization.} Strong guidance can bias samples toward conservative structures, potentially reducing diversity or favoring common motifs.
\end{itemize}

\subsection{Sampling efficiency and scalability}
\begin{itemize}
    \item \textbf{Compute overhead.} Coupled ODE integration and guidance evaluation add runtime relative to simpler generators.
    \item \textbf{Step-size / solver dependence.} Sample quality and constraint satisfaction may depend on solver choice, tolerances, and step count.
    \item \textbf{Scaling to larger systems.} Extending to larger molecules or macromolecular settings may require more efficient guidance, amortized constraints, or hierarchical solvers.
\end{itemize}

\subsection{Benchmark and evaluation caveats}
\begin{itemize}
    \item \textbf{GEOM-DRUGS pipeline issues.} Reported absolute metrics on GEOM-DRUGS can be affected by known preprocessing/valency-table and bond-order computation issues, as well as force-field inconsistencies \cite{nikitin2025geomdrugsrevisited}.
    \item \textbf{Metric dependence on bond perception.} Many validity/uniqueness/novelty statistics depend on the procedure used to infer bond orders from 3D coordinates; different toolchains may yield different outcomes.
    \item \textbf{Generality beyond drug-like space.} Current experiments focus on small drug-like molecules; performance on unusual chemistries, charged/metal-containing compounds, or biomolecules remains to be established.
\end{itemize}
\end{document}